%% file: paper.tex
\documentclass[12pt]{article}

\usepackage{fancyhdr,amsmath,amsthm,amssymb,bm} 
\usepackage{graphicx,caption}
\usepackage{float}
\usepackage{bm}
\usepackage{bbm}
\usepackage{amsfonts,verbatim,url}
\usepackage{dsfont}
\usepackage[top=1in, bottom=1in, left=1in, right=1in]{geometry}
\usepackage[square,numbers]{natbib}
\usepackage{color}
\usepackage{setspace} 
\usepackage{algorithm}
\usepackage{algorithmic}
\usepackage{blindtext}
\usepackage{verbatim}
\usepackage{mkolar_definitions}
\usepackage{textcomp}
\usepackage{authblk}
\usepackage{diagbox}

\usepackage{mathtools}
\mathtoolsset{showonlyrefs}

\usepackage[colorlinks=true,
            linkcolor=blue,
            urlcolor=blue,
            citecolor=blue]{hyperref}

\let\tilde\widetilde
\let\hat\widehat

\graphicspath{{fig/}}

\begin{document}
\title{Learning Influence-Receptivity Network Structure  \\ with Guarantee}

\author{Ming Yu\thanks{Booth School of Business, The University of Chicago. Email: \href{mailto:mingyu@chicagobooth.edu}{mingyu@chicagobooth.edu} } , 
Varun Gupta\thanks{Booth School of Business, The University of Chicago. Email: \href{mailto:varun.gupta@chicagobooth.edu}{varun.gupta@chicagobooth.edu}} , 
and Mladen Kolar\thanks{Booth School of Business, The University of Chicago. Email: \href{mailto:mladen.kolar@chicagobooth.edu}{mladen.kolar@chicagobooth.edu}}}


\date{}
\maketitle

\input{Abstract}

\input{Introduction}

\input{Model}

\input{Optimization}

\input{Theoretical}

\input{Joint}

\input{Synthetic}

\input{Real}

\input{Conclusion}

\section*{Acknowledgments}

This work is partially supported by an IBM Corporation Faculty
Research Fund at the University of Chicago Booth School of
Business. This work was completed in part with resources provided by
the University of Chicago Research Computing Center.

\newpage
\bibliographystyle{plain}
\bibliography{paper} 

\newpage
\appendix
\input{Appendix}

\end{document}

%% file: Abstract.tex
\begin{abstract}
\label{sec:abstract}
Traditional works on community detection from observations of
information cascade assume that a single adjacency matrix parametrizes
all the observed cascades.  However, in reality the connection
structure usually does not stay the same across cascades.  For
example, different people have different topics of interest, therefore
the connection structure depends on the information/topic content of
the cascade.  In this paper we consider the case where we observe a
sequence of noisy adjacency matrices triggered by information/event
with different topic distributions.  We propose a novel latent model
using the intuition that a connection is more likely to exist between
two nodes if they are interested in similar topics, which are common
with the information/event.  Specifically, we endow each node with two
node-topic vectors: an influence vector that measures how 
influential/authoritative they are on each topic; and a receptivity
vector that measures how  receptive/susceptible they are to each
topic.  We show how these two node-topic structures can be estimated
from observed adjacency matrices with theoretical guarantee on
estimation error, in cases where the topic distributions of the
information/event are known, as well as when they are unknown.
Experiments on synthetic and real data demonstrate the effectiveness
of our model and superior performance compared to state-of-the-art
methods.
\end{abstract}

%% file: Introduction.tex
\section{Introduction}
\label{sec:introduction}

Uncovering latent network structure is an important research area in
network model and has a long history \cite{wasserman1994social,
  burt2000network}.  For a $p$ node network, traditional approaches
usually assume a single $p \times p$ adjacency matrix, either binary
or real-valued, that quantifies the connection intensity between
nodes, and aim to learn the community structure from it.  For example,
in Stochastic Block Model (SBM) \cite{holland1983stochastic} we assume
that nodes within a group have an edge with each other with
probability $p_0$ while nodes across groups have an edge with
probability $q_0$ where $p_0 > q_0$.  In information diffusion we
observe the propagation of information among nodes and aim to recover
the underlying connections between nodes
\cite{rodriguez2011uncovering, gomez2015estimating,
  gomez2010inferring}.  In time-varying networks we allow the
connections and parameters to change over time
\cite{kolar2010estimating, ahmed2009recovering}.  In this paper, we
consider the case where we have a sequence of information/event/collaboration with
different topics, and we observe a noisy adjacency matrix for each of
them.  The connection between nodes varies under each topic
distribution and this cannot be captured by only one adjacency
matrix.  For example, each researcher has her own research interests
and would collaborate with others only on the areas they are both
interested in. Specifically, suppose researcher 1 is interested in
computational biology and information theory; researcher 2 is
interested in computational biology and nonparametric statistics;
researcher 3 is interested in information theory only. Then if
researcher 1 wants to work on computational biology, she would
collaborate with researcher 2; while if the topic is on information
theory, then she would collaborate with researcher 3.  As another
example, suppose student 1 is interested in music and sports while
student 2 is interested in music and chess. If the topic of a
University event is music, then there will be an edge between these
two students; however, if the topic of the event is sports or chess,
then there would not be an edge between them.

Intuitively, for a specific information/event/collaboration, there
will be an edge between two nodes if and only if they are both
interested in the topic of this information/event/collaboration.  In
this paper we model this intuition by giving each node two node-topic
vectors: one \emph{influence} vector (how authoritative they are on
each topic) and one \emph{receptivity} vector (how susceptible they
are on each topic).  In addition, each information/event/collaboration
is associated with a distribution on topics.  The influence and
receptivity vectors are fixed but different topic distributions result
in different adjacency matrices among nodes.  In this paper we
consider both cases where the topic distribution may or may not be known,
and provide algorithms to estimate the node-topic structure with
theoretical guarantees on estimation error.  In particular, we show
that our algorithm converges to the true values up to statistical
error.  Our node-topic structure is easier to interpret than a large
adjacency matrix among nodes, and the result can be used to make
targeted advertising or recommendation systems.

\vspace{-2mm}

\paragraph{Notation}

In this paper we use $p$ to denote the number of nodes in the network; we assume there are $K$ topics in total, and we observe $n$ adjacency matrices under different topic distributions. 
We use subscript $i \in \{1,\ldots,n\}$ to index samples/observations; subscript $j,\ell \in\{1, \ldots, p\}$ to index nodes; and subscript $k \in \{1,\ldots, K\}$ to index topic. 
For any matrix $A$, we use $\|A\|_0 = |(j,k): A_{jk} \neq 0|$ to denote the number of nonzero elements of $A$. Also, for any $d$, $I_d$ is the identity matrix with dimension $d$.




%% file: Model.tex
\section{Model}
\label{sec:model}

Our model to capture the node-topic structure in networks is built on
the intuition that, for a specific information/event/collaboration,
there would be an edge between two nodes if they are interested in
similar topics, which are also common with that of the
information/event/collaboration. Furthermore, the connection is
\emph{directed} where an edge from node 1 to node 2 is more likely to
exist if node 1 is influential/authoritative in the topic, and node 2
is receptive/susceptible to the topic.  For example, an eminent
professor would have a large influence value (but maybe a small
receptivity value) on his/her research area, while a high-producing,
young researcher would have a large receptivity value (but maybe a
small influence value) on his/her research area. Note that the notion
of ``topic'' can be very general.  For example it can be different
immune systems: different people have different kinds of immune
systems, and a disease is more likely to propagate between people with
similar and specific immune system.

Our node-topic structure is parametrized by two matrices
$B_1, B_2 \in \mathbb{R}^{p \times K}$. The matrix $B_1$ measures how
much a node can infect others (the \emph{influence} matrix) and the
matrix $B_2$ measures how much a node can be infected by others (the
\emph{receptivity} matrix).  We use $b_{jk}^1$ and $b_{jk}^2$ to
denote the elements of $B_1$ and $B_2$, respectively. Specifically,
$b_{jk}^1$ measures how influential node $j$ is on topic $k$, and
$b_{jk}^2$ measures how receptive node $j$ is on topic $k$.  We use
$b_{k}^1$ and $b_{k}^2$ to denote the columns of $B_1$ and $B_2$,
respectively.

Each observation $i$ is associated with a topic distribution
$m_i = (m_{i1}, ..., m_{iK})$ on the $K$ topics satisfying
$m_{i1}, ..., m_{iK} \geq 0$ and $m_{i1} + ... + m_{iK} = 1$.  The
choice of $K$ can be heuristic and pre-specified or alternatively can
be decided by methods such as in \cite{hsu2016online} which learn the
distribution over the number of topics.  For each observation $i$, the
true adjacency matrix is given by
\begin{equation}
\big(x^*_i\big)_{j\ell} = \sum_{k=1}^K b_{jk}^1 \cdot m_{ik} \cdot  b_{\ell k}^2,
\label{eq:xistar}
\end{equation}
or in matrix form,
\begin{equation}
X^*_i = B_1 \cdot M_i \cdot B_2^\top,
\label{eq:Xistar}
\end{equation}
where $M_i$ is a diagonal matrix 
\begin{equation}
M_i = \diag(m_{i1}, m_{i2}, ..., m_{iK}).
\end{equation}

The interpretation of the model is straightforward from
\eqref{eq:xistar}.  For an observation $i$ on topic $k$, there will be
an edge $j \to \ell$ if and only if node $j$ tends to infect others on
topic $k$ (large $b_{jk}^1$) and node $\ell$ tends to be infected by
others on topic $k$ (large $b_{\ell k}^2$). This intuition applies to
each topic $k$ and the final value is the summation over all the $K$
topics.

If we do not consider self connections, we can zero out the diagonal
elements and get
\begin{equation}
X_i^* = B_1 M_i B_2^\top - \text{diag}(B_1 M_i  B_2^\top).
\label{eq:Xistar_diag}
\end{equation}

For notational simplicity, we still stick to
\eqref{eq:Xistar} for the definition of $X_i^*$ in the subsequent
sections. The data consists of $n$ observations $\{X_i\}_{i=1}^n$ satisfying
\begin{equation}
\label{eq:observe_Xi}
X_i = X_i^* + E_i,
\end{equation}
where the noise term $E_i$ are mean 0 and independent across $i$. They are not necessarily identically
distributed and can follow an unstructured distribution. The
observations $X_i$ can be either real-valued or binary. For binary
observations we are interested in the existence of a connection only,
while for real-valued observation we are also interested in how strong
the connection is, i.e. larger values indicate stronger connections.

\paragraph{Related Works}

There is a vast literature on uncovering latent network structures.
The most common and basic model is the Stochastic block model (SBM)
\cite{holland1983stochastic} where connections are assumed to be dense
within group and are sparse across groups.  The exact recovery of SBM
can be solved using maximum likelihood method but is NP-hard.
Many practical algorithms have been proposed for SBM such as Modularity
method, EM algorithm, Spectral clustering, etc
\cite{bickel2009nonparametric, karrer2010message, rohe2011spectral,
  nowicki2001estimation, snijders1997estimation,wang1987stochastic}.
Many variants and extensions of SBM have also been developed to
better fit real world network structures, including Degree-corrected
block model (DCBM) \cite{karrer2011stochastic}, Mixed membership
stochastic block models (MMSB) \cite{airoldi2008mixed}, Degree
Corrected Mixed Membership (DCMM) model \cite{jin2017estimating}, etc.
Other models include information diffusion
\cite{rodriguez2011uncovering, gomez2015estimating,
  gomez2010inferring, zhou2013learning}, time-varying networks
\cite{kolar2010estimating, ahmed2009recovering}, 
conjunctive Boolean networks \cite{gao2018stability, jarrah2010dynamics, gao2018controllability}, 
graphical models \cite{Lauritzen1996Graphical, Barber2015ROCKET, yu2016statistical}, 
buyer-seller networks
\cite{kranton2001theory, wilson1995integrated, walter2001value},
etc. \cite{hoff2002latent} and \cite{hoff2009multiplicative} assume a
``logistic'' model based on covariates to determine whether an
edge exists or not.  However, most of the existing work focuses on a
single adjacency matrix and ignores the node-topic structure.
In \cite{Yu2017AnIM} the authors propose a node-topic model for information diffusion problem, but it requires the topic distribution to be known and lacks theoretical guarantees.

In \cite{airoldi2013stochastic} the authors study multiple adjacency
matrices but it still falls into the SBM framework. The number of
blocks need to be predefined (the performance is sensitive to this
value) and the output is the block information. As a contrast, our
model outputs the (numeric) influence-receptivity information for each
node and these nodes do not need to form blocks. Also, their work does
not utilize the topic information.

In terms of topic-based network inference, a closely related work is
\cite{du2013uncover} where the authors use $K$ adjacency matrices to
describe the network structure. However it ignores the node-topic
structure and can only deal with the case where the topic
distributions are known, while our method is able to learn the topic
distribution and the network structure simultaneously.
In Section \ref{sec:synthetic} and \ref{sec:real} we show that our method outperforms this model on both synthetic and real dataset.

Another closely related work is \cite{chen2017fast} where the authors
propose the graph embedding model which also gives each node two $K$
dimensional ``embedding'' vector. However, our model is different in
the following senses: 1. The topic information of our model is easier
to interpret than the ``embedding'' vectors.  The whole framework of
our model is more interpretable: we know all the topics information
and the topics of interest for each node.  2. We provide a generative
model and thorough theoretical result (error analysis).  3. The graph
embedding model focuses on only one observation while our model
focuses on $n$ observations with each observation having a different
topic distribution. In our model, the influence and receptivity
vectors interact with topic information, while the graph embedding
model cannot deal with that.

If we add up all the adjacency matrices $X_i$ to a single matrix $X$, then it is similar to the mixed membership stochastic block model (MMSB \cite{airoldi2008mixed}) where $X=B*M*B^\top$ and $M$ can be non-diagonal. Compared to MMSB, our model allows for asymmetry by considering ``influence'' and ``receptivity''; our model considers that information with a different topic can have different adjacency matrices; also, our model can be used to predict a future adjacency matrix given the topics. Finally, when we have $n$ adjacency matrices, it is usually better to analyze them individually instead of adding them up, which may lead to information loss.

%% file: Optimization.tex
\section{Optimization}
\label{sec:optimization}

In this paper we consider the loss function
\begin{equation}
\label{eq:loss_B1B2}
f(B_1,B_2) = \frac{1}{2n} \sum_{i=1}^n \big\|X_i - B_1 M_i B_2^\top \big\|_F^2.
\end{equation}
Using the notation $B_1 = [b^1_1, ..., b^1_K]$ and
$B_2 = [b^2_1, ..., b^2_K]$, we can rewrite \eqref{eq:Xistar} as
\begin{equation}
X_i^* = B_1  M_i  B_2^\top = \sum_{k=1}^K m_{ik} \cdot b^1_k {b^2_k}^\top.
\label{eq:Xistar_rewrite}
\end{equation}

Denote $\Theta_k = b^1_k {b^2_k}^\top$; with some abuse
of notation we can rewrite the loss function \eqref{eq:loss_B1B2} as
\begin{equation}
\label{eq:loss_Theta}
f(\Theta) = f(\Theta_1, ..., \Theta_K) = \frac{1}{2n} \sum_{i=1}^n \Big\|X_i - \sum_{k=1}^K m_{ik} \cdot \Theta_k \Big\|_F^2 .
\end{equation}
From \eqref{eq:loss_Theta} we can see that solving for $B_1,B_2$ is
equivalent to solving for rank-1 matrix factorization problem on
$\Theta_k$.  This model is therefore not identifiable on $B_1$ and
$B_2$, since if we multiply column $k$ of $B_1$ by some scalar
$\gamma$ and multiply column $k$ of $B_2$ by $1/\gamma$, the matrix
$X_i^*$ remains unchanged for any $i$, since $b^1_k {b^2_k}^\top$ does
not change. Hence the loss function also remains unchanged. Therefore
we need an additional regularization term to ensure a unique solution.  
To address this issue, we propose the following two alternative regularization terms.
\begin{enumerate}
\item The first regularization term is an $L_1$ penalty on $B_1$ and
  $B_2$. We define the following norm
\begin{equation}
  \|B_1 + B_2\|_{1,1} \triangleq \lambda \cdot \Big( \sum_{j, k} b^1_{jk} + b^2_{jk} \Big) ,
\end{equation}
where $\lambda$ is a tuning parameter. To see why this penalty ensures
unique solution, we focus on column $k$ only. The term we want to
minimize is
\begin{equation}
\gamma \cdot \|b^1_k\|_1 + \frac{1}{\gamma} \|b^2_k\|_1.
\label{eq:column}
\end{equation}
In order to minimize \eqref{eq:column} we should select $\gamma$ such
that the two terms in \eqref{eq:column} are equal. In other words, the
column sums of $B_1$ and $B_2$ are equal.
\item The second regularization term is borrowed from matrix
  factorization literature defined as
\begin{equation}
\label{eq:guv}
g(B_1,B_2) = \frac \lambda 2 \cdot \sum_{k=1}^K \Big( \big\|b^1_k\big\|_2^2 - \big\|b^2_k \big\|_2^2 \Big)^2.
\end{equation}
This regularization term forces the 2-norm of each column of $B_1$ and
$B_2$ to be the same. 
\end{enumerate}
Both regularization terms force the columns of $B_1$ and $B_2$ to be
balanced.  
Intuitively, this means that, for each topic $k$, the total
magnitudes of ``influence'' and ``receptivity'' are the same. This
acts like a conservation law that the total amount of output should be
equal to the total amount of input.
At the minimizer, this regularization term is 0,
and therefore we can pick any $\lambda > 0$.

The first regularization term introduces bias, but it encourages
sparse solution; the second regularization term does not introduce
bias, but we need an additional hard thresholding step to get
sparsity.  Experimentally, both regularizations work; theoretically,
the loss function \eqref{eq:loss_B1B2} is nonconvex in $B_1$ and
$B_2$, hence proving theoretical results is much harder for the first
regularization term. Therefore our theoretical results focus on the
second alternative proposed above.  
The final optimization problem is
given by
\begin{eqnarray}
\begin{aligned}
& \text{minimize} \quad \frac{1}{2n} \sum_{i=1}^n \big\|X_i - B_1 M_i B_2^\top \big\|_F^2  + \frac \lambda 2 \cdot \sum_{k=1}^K \Big( \big\|b^1_k\big\|_2^2 - \big\|b^2_k \big\|_2^2 \Big)^2   \\
& \text{subject to} \quad B_1, B_2 \geq 0 
\label{eq:optimization}
\end{aligned}
\end{eqnarray}

\paragraph{Initialization.} We initialize by solving the convex
relaxation problem \eqref{eq:loss_Theta} without the rank-1 constraint
on $\Theta_k$, and apply rank-1 SVD on estimated $\hat \Theta_k$,
i.e., we keep only the largest singular value:
$[u_k, s_k, v_k] = {\text{rank-}}1 { \text{ SVD of }} \hat
\Theta_k$.
The initialization is given by
$B_1^{(0)} = [u_1s_1^{1/2}, ..., u_Ks_K^{1/2}]$ and
$B_2^{(0)} = [v_1s_1^{1/2}, ..., v_Ks_K^{1/2}]$.  Being a convex
relaxation, we can find the global minimum $\hat \Theta_k$ of problem
\eqref{eq:loss_Theta} by using gradient descent algorithm.

\paragraph{Algorithm.}
After the initialization, we alternately apply proximal gradient
method \cite{parikh2014proximal} on $B_1$ and $B_2$ until convergence.
In practice, each node would be interested in only a few topics and
hence we would expect $B_1$ and $B_2$ to be sparse.  To encourage
sparsity we need an additional hard thresholding step on $B_1$ and
$B_2$.  The overall procedure is given in Algorithm
\ref{algo:B1B2}. The operation $\text{Hard}(B,s)$ keeps the largest
$s$ elements of $B$ and zeros out others; the operation $[B]_+$ keeps
all positive values and zeros out others.


\begin{algorithm}[tb]
   \caption{Alternating proximal gradient descent}
   \label{algo:B1B2}
\begin{algorithmic}
   \STATE {\bfseries Initialize $B_1^{(0)}$, $B_2^{(0)}$}
   \FOR{$t = 1, ..., T$}
   \STATE $B_1^{(t+0.5)} = \Big[B_1^{(t)} - \eta  \nabla_{B_1} f\big(B_1^{(t)}, B_2^{(t)}\big) - \eta  \nabla_{B_1} g\big(B_1^{(t)}, B_2^{(t)}\big)\Big]_+$
   \STATE $B_1^{(t+1)} = \text{Hard}\big(B_1^{(t+0.5)}, s\big)$
   \STATE $B_2^{(t+0.5)} = \Big[B_2^{(t)} - \eta \cdot \nabla_{B_2} f\big(B_1^{(t)}, B_2^{(t)}\big) - \eta \cdot \nabla_{B_2} g\big(B_1^{(t)}, B_2^{(t)}\big)\Big]_+$
   \STATE $B_2^{(t+1)} = \text{Hard}\big(B_2^{(t+0.5)}, s\big)$
   \ENDFOR
   
\end{algorithmic}
\end{algorithm}

%% file: Theoretical.tex
\section{Theoretical result}
\label{sec:theoretical}

In this section we derive the theoretical results for our algorithm. We denote $B_1^*$ and $B_2^*$ as the true value and $\Theta_k^* = {b^1_k}^* {b^2_k}^{*\top}$ as the corresponding true rank-1 matrices. In this section we assume the topic distribution $M_i$ is known. The case where $M_i$ is unknown is considered in Section \ref{sec:joint}. 
All the detailed proofs are relegated to the Appendix. We start by stating the following two mild assumptions on the parameters of the problem.

\paragraph{Topic Condition (TC).} Denote the Hessian matrix on $\Theta$ as
\begin{equation}
\begin{small}
H_{\Theta} = 
\frac{1}{n}
\begin{bmatrix}
\vspace{1mm}
\sum_i m_{i1}^2 & \sum_i m_{i1}m_{i2} & \dots & \sum_i m_{i1}m_{iK} \\
\sum_i m_{i1}m_{i2}  &  \sum_i m_{i2}^2 &  \dots & \sum_i m_{i2}m_{iK} \\
\vdots & \vdots & \ddots & \vdots \\
\sum_i m_{i1}m_{iK} & \sum_i m_{i2}m_{iK} & \dots & \sum_i m_{iK}^2
\end{bmatrix}.
\end{small}
\end{equation}
We require $H_{\Theta} \succeq \mu_\Theta \cdot I_K$ for some constant $\mu_\Theta > 0$.

Intuitively, this condition requires that, the correlation among topic
distributions in the $n$
observations cannot be too large.  This makes sense because if several
topics are highly correlated with each other among the
$n$
observations, then clearly we cannot distinguish them. 
If we vectorize each $\Theta_k$, the Hessian matrix of $f(\Theta)$ with respect to $\Theta$ is a $p^2K$ by $p^2K$ matrix and it can be shown that this Hessian matrix is given by $H_{\Theta} \otimes I_{p^2}$ where $\otimes$ is the Kronecker product. 
With this
condition, the objective function \eqref{eq:loss_Theta} is strongly
convex in $\Theta$.

An immediate corollary of this condition is that the diagonal elements
of $H_{\Theta}$ must be at least $\mu_\Theta$, i.e., for each topic
$k$, we have $\frac 1n \sum_{i=1}^n m_{ik}^2 \geq \mu_\Theta$. This
means that at least a constant proportion of the observed data should
focus on this topic. The necessity of this condition is also
intuitive: if we only get tiny amount of data on some topic, then we
cannot expect to recover the structure for that topic accurately.

\paragraph{Sparsity Condition (SC).} Both $B_1^*$ and $B_2^*$ are sparse: $\|B_1^*\|_0 = \|B_2^*\|_0 = s^*$. (We use a single $s^*$ for notational simplicity, but is not required).

\paragraph{Subspace distance.} For matrix factorization problems, it
is common to measure the subspace distance because the factorization
$\Theta_k = b^1_k {b^2_k}^\top$ is not unique. Here since we know that
$\Theta_k$ are exactly rank-1 and we have non-negativity constraints
on $B_1, B_2$, we would not suffer from rotation issue (the only way
to rotate scalar is $\pm1$, but with non-negative constraint, $-1$ is
impossible).  Therefore the subspace distance between $B = [B_1, B_2]$
and $B^* = [B_1^*, B_2^*] $ is just defined as
\begin{equation}
\begin{aligned}
d^2(B,B^*)
&= \min_{o_k \in \{\pm1\}}\sum_{k=1}^K  \|b_k^1 - {b_k^1}^*o_k\|_2^2 + \|b_k^2 - {b_k^2}^*o_k\|_2^2  = \|B_1 - B_1^*\|_F^2 + \|B_2 - B_2^*\|_F^2.
\end{aligned}
\end{equation}

\paragraph{Statistical error.} Denote
\begin{equation}
\begin{aligned}
\Omega &= \big\{ \Delta: \Delta = [\Delta_1, ..., \Delta_K] \in \RR^{pK \times p}, {\rm rank}(\Delta_k) = 2,  \|\Delta_k\|_0 = s, \|\Delta\|_F = 1 \big\}.
\end{aligned}
\end{equation}

The statistical error on $\Theta$ is defined as
\begin{equation}
\begin{aligned}
\label{eq:def_stat_error}
e_{\text{stat}, \Theta} &= \sup_{\Delta \in \Omega} \, \big\langle \nabla f_{\Theta}(\Theta^*), \Delta \big\rangle = \sup_{\Delta \in \Omega} \, \sum_{k=1}^K \Big\langle \frac 1n \sum_{i=1}^n E_i \cdot m_{ik}, \Delta_k \Big\rangle.
\end{aligned}
\end{equation}
where $E_i$ is the error matrix in \eqref{eq:observe_Xi} and $\langle A, B \rangle=\tr(A^\top B)$ is the matrix inner product. 
Intuitively, this statistical error measures how much accuracy we can
expect for the estimator. If we are within
$c \cdot e_{\text{stat}}$ distance with the true value, then we are
already optimal.

The statistical error depends on the sparsity level $s$. In practice, $s$ is a hyperparameter and one can choose it as a relatively large value to avoid missing true nonzero values. If $s$ is too large, then we include too many false positive edges. This usually does not affect performance too much, since these false positive edges tend to have small values. However, we lose some sparsity and hence interpretability. If we further assume that each node is interested in at least one but not most of the topics, then we have $s=O(p)$ and we can choose $s=c \cdot p$ where $c$ can be a small constant. In this way, the effect of choosing $s$ is minimal.

In this way we transform the original problem to a standard matrix
factorization problem with $K$ rank-1 matrices
$\Theta_1, \ldots,
\Theta_K$. 
A function $f(\cdot)$ is termed to be strongly convex and smooth if
there exist constant $\mu$ and $L$ such that
\begin{equation}
\begin{aligned}
\frac \mu 2 \big\| Y - X \big\|_F^2 \leq f(Y) - f(X) & - \langle \nabla f(X), Y - X \rangle \leq \frac L 2 \big\| Y - X \big\|_F^2. 
\end{aligned}
\end{equation}

The objective function \eqref{eq:loss_Theta} is
strongly convex and smooth in $\Theta$.  Since the loss function
\eqref{eq:loss_Theta} is quadratic on each $\Theta_k$, it is easy to
see that the conditions are equivalent to
$\mu \cdot I_K \preceq H_\Theta \preceq L \cdot I_K$.
The lower bound is satisfied according to assumption (TC) with
$\mu = \mu_\Theta$, and the upper bound is trivially satisfied with
$L = L_{\Theta}=1$. Therefore we see that the objective function
\eqref{eq:loss_Theta} is strongly convex and smooth in $\Theta$.
The following lemma quantifies the accuracy of the initialization.

\begin{lemma}
\label{lemma:init}
Suppose $\hat\Theta = (\hat \Theta_1,\ldots, \hat \Theta_K)$ are the global minimum of the convex relaxation \eqref{eq:loss_Theta}, then we have
\begin{equation}
\sum_{k=1}^K \|\Theta_k^* - \hat \Theta_k \|_F^2 \leq \frac{2}{\mu_\Theta} \big\| \nabla f(\Theta^*) \big\|_F.
\end{equation}
\end{lemma}

The bound we obtain from Lemma \ref{lemma:init} scales
with $n^{-1/2}$ and therefore can be small as long as we have enough
samples.  We are then ready for our main theorem. The following
Theorem \ref{theorem:known} shows that the iterates of Algorithm
\ref{algo:B1B2} converge linearly up to statistical error.


\begin{theorem}
\label{theorem:known}
Suppose conditions (SC) and (TC) hold. We set the sparsity level
$s = cs^*$. If the step size $\eta$ satisfies
\begin{equation}
\label{eq:eta_constant}
\eta \leq \frac{1}{16\|B^{(0)}\|_2^2} \cdot \min\Big\{\frac{1}{2(\mu_{\Theta} +L_{\Theta} )}, 1\Big\}, 
\end{equation}
then for large enough $n$, after $T$ iterations, we have
\begin{equation}
\label{eq:linear_convergence}
d^2 \big( B^{(T)}, B^* \big) \leq \beta^T d^2 \big( B^{(0)}, B^* \big) + C\cdot e^2_{\rm{stat}, \Theta},
\end{equation}
for some constant $\beta < 1$ and constant $C$.
\end{theorem}

\begin{remark}
  Although we focus on the simplest loss function
  \eqref{eq:loss_B1B2}, our analysis works for any general loss
  functions $f(B_1MB_2^\top)$, as long as the initialization is
  good and the (restricted) strongly convex and smoothness
  conditions are satisfied. See \cite{yu2018recovery} for more details.
\end{remark}

\begin{remark}
\label{remark:complexity}
For time complexity of Algorithm \ref{algo:B1B2}, calculating the gradient takes $O(p^2K)$ time and hence taking average over all samples takes $O(np^2K)$ time. The initialization step involves SVD; 
but we do not need to obtain the full decomposition since for each $\Theta_k$ we only need the singular vector corresponding to the largest singular value.
Finally, the number of iteration $T$ is such that $\beta^T$ has the same order with the statistical error, which gives $T<O(\log n)$.
\end{remark}

%% file: Joint.tex
\section{Learning network and topic distributions jointly}
\label{sec:joint}

So far we have assumed that the topic distributions $m_i$ for each sample $i$ are
given and fixed.
However, sometimes we do not have such information.  In this case we
need to learn the topic distributions and the network structure
simultaneously.

We denote $m_i^*$ as the true topic distribution of observation $i$
and $M = [m_1, ..., m_n]$ is the stack of all the topic distributions.
The algorithm for joint learning is simply alternating minimization on
$B_1,B_2$ and $M$. For fixed $M$, the optimization on $B_1, B_2$ is
the same as before, and can be solved using Algorithm
\ref{algo:B1B2}. For fixed $B_1,B_2$, it is straightforward to see
that the optimization on $M$ is separable for each $i$. For each $i$,
we solve the following optimization problem to estimate
$M_i = \diag(m_i)$:
\begin{equation}
\begin{aligned}
\label{eq:optimization_M}
& \text{minimize} \quad  \big\|X_i - B_1 M_i B_2^\top \big\|_F^2   \\
& \text{subject to} \quad m_i \geq 0, 1^\top \cdot m_i = 1
\end{aligned}
\end{equation}

This problem is convex in $M_i$ and can be easily solved using
projected gradient descent. Namely in each iteration we do gradient
descent on $M_i$ and then project to the simplex. The overall
procedure is summarized in Algorithm \ref{algo:joint}.  With some
abuse of notation we write
\begin{equation}
\label{eq:loss_Theta_M}
f(\Theta, M) = \frac{1}{2n} \sum_{i=1}^n \Big\|X_i - \sum_{k=1}^K m_{ik} \cdot \Theta_k \Big\|_F^2.
\end{equation}

Besides the scaling issue mentioned in Section \ref{sec:optimization},
the problem now is identifiable only up to permutation of the position
of the topics. However we can always permute $M^*$ to match the
permutation obtained in $M$. From now on we assume that these two
permutations match and ignore the permutation issue.  The statistical
error on $M$ is defined as
\begin{equation}
\begin{aligned}
e_{\text{stat}, M}^2 &= \sum_{i,k} \Big[ \nabla_{m_{ik}} f(\Theta^*, M^*) \Big]^2 = \frac{1}{n} \sum_{i=1}^n \sum_{k=1}^K  \langle E_i, \Theta_k^* \rangle ^2.
\end{aligned}
\end{equation}

The problem is much harder with unknown topic distribution.
Similar to condition (TC), we need the following assumption on the
Hessian matrix on $M$.

\begin{algorithm}[tb]
   \caption{Learning network structure and topic distributions jointly}
   \label{algo:joint}
\begin{algorithmic}
   \STATE {\bfseries Initialize $B_1$, $B_2$}
   \WHILE{$tolerance > \epsilon$}
   \STATE Optimize $M$ according to \eqref{eq:optimization_M} using projected gradient descent.
   \STATE Optimize $B_1, B_2$ according to Algorithm \ref{algo:B1B2}
   \ENDWHILE
\end{algorithmic}
\end{algorithm}

\paragraph{Diffusion Condition (DC).} Denote the Hessian matrix on $M$ as 
\begin{equation}
H_{M} = 
\begin{bmatrix}
\vspace{1mm}
\langle \Theta_1^*, \Theta_1^* \rangle & \langle \Theta_1^*, \Theta_2^* \rangle & \dots & \langle \Theta_1^*, \Theta_K^* \rangle \\
\langle \Theta_2^*, \Theta_1^* \rangle & \langle \Theta_2^*, \Theta_2^* \rangle &  \dots & \langle \Theta_2^*, \Theta_K^* \rangle \\
\vdots & \vdots & \ddots & \vdots \\
\langle \Theta_K^*, \Theta_1^* \rangle & \langle \Theta_K^*, \Theta_2^* \rangle & \dots & \langle \Theta_K^*, \Theta_K^* \rangle
\end{bmatrix},
\end{equation}
where $\langle A_1, A_2\rangle=\tr(A_1^\top A_2)$ is the inner product
of matrices $A_1, A_2$. We require that
$H_{M} \succeq \mu_M \cdot I_K$ for some constant $\mu_M > 0$.

With this condition, the objective function \eqref{eq:loss_Theta} is
strongly convex in $M$. The intuition is similar as in condition
(TC). We require that $\Theta_k$ can be distinguished from each other.

\paragraph{Initialization.}
Define $\overline X$, $\overline X^*$, $\overline E$ as the sample mean of $X_i$, $X_i^*$, $E_i$, respectively. It is clear that $\overline X = \overline X^* + \overline E$. 
We then do rank-$K$ svd on $\overline X$ and obtain $ [\tilde U, \tilde S, \tilde V] = {\text{rank-}}K { \text{ svd of }} \overline X$. 
We denote $\tilde X = \tilde U\tilde S\tilde V^\top = \sum_{k=1}^K \tilde \sigma_k \tilde u_k \tilde v_k^\top$ and we initialize with
\begin{equation}
\label{eq:joint_init_rule}
\Theta_k^{(0)} = K \cdot \tilde \sigma_k \tilde u_k \tilde v_k^\top.
\end{equation}
To see why this initialization works, we first build intuition for the easiest case, where $E_i = 0$ for each $i$, $\frac{1}{n} \sum_{i=1}^n m_{ik}^* = \frac{1}{K}$ for each $k$, and the columns of $B_1^*$ and $B_2^*$ are orthogonal. In this case it is easy to see that $\overline X = \overline X^* = \sum_{k=1}^K \frac{1}{K} \Theta_k^*$. 
Note that this expression in a singular value decomposition of $\overline X^*$ since we have $\Theta_k^* = {b^1_k}^* {b^2_k}^{*\top}$ and the columns $\{ {b^1_k}^* \}_{k=1}^K$ and columns $\{ {b^2_k}^* \}_{k=1}^K$ are orthogonal. 
Now that $\overline X$ is exactly rank $K$, the best rank $K$ approximation would be itself, i.e., $\overline X = \tilde X = \sum_{k=1}^K \tilde \sigma_k \tilde u_k \tilde v_k^\top$. 
By the uniqueness of singular value decomposition, as long as the singular values are distinct, we have (up to permutation) $\frac{1}{K} \Theta_k^* = \tilde \sigma_k \tilde u_k \tilde v_k^\top$ and therefore $\Theta_k^* = K \cdot \tilde \sigma_k \tilde u_k \tilde v_k^\top$. This is exactly what we want to estimate. 

\vspace{2mm}
In order to show this is a reasonable initialization, we impose the following condition.

\paragraph{Orthogonality Condition (OC).} Let $B_1^* = Q_1R_1$ and $B_2^* = Q_2R_2$ be the QR decomposition of $B_1^*$ and $B_2^*$, respectively. Denote $A^*$ as a diagonal matrix with diagonal elements $\frac{1}{n} \sum_{i=1}^n m_{ik}^*$. Denote $R_1 A^* R_2^\top = A_{\rm diag} + A_{\rm off}$ where $A_{\rm diag}$ captures the diagonal elements and $A_{\rm off}$ captures the off-diagonal elements. We require that $\|A_{\rm off}\|_F \leq \rho_0$ for some constant $\rho_0$. Moreover, we require that $\frac{1}{n} \sum_{i=1}^n m_{ik}^* \leq \eta / K$ for some $\eta$. 

This condition requires that $B_1^*$ and $B_2^*$ are not too far away from orthogonal matrix, so that when doing the QR rotation, the off diagonal values of $R_1$ and $R_2$ are not too large. 
The condition $\frac{1}{n} \sum_{i=1}^n m_{ik}^* \leq \eta / K$ is trivially satisfied with $\eta = K$. However, in general $\eta$ is usually a constant that does not scale with $K$, meaning that the topic distribution among the $n$ observations is more like evenly distributed than dominated by a few topics.

It is useful to point out that the condition (OC) is for this specific initialization method only. 
Since we are doing singular value decomposition, we end up with orthogonal vectors so we require that $B_1^*$ and $B_2^*$ are not too far away from orthogonal; since we do not know the value $\frac{1}{n} \sum_{i=1}^n m_{ik}^*$ and use $1/K$ to approximate, we require that topics are not far away from evenly distributed so that this approximation is reasonable. 
In practice we can also use other initialization methods, for example we can do alternating gradient descent on $\Theta$ and $M$ based on the objective function \eqref{eq:loss_Theta_M}. This method also works reasonably well in practice.

\vspace{3mm}
The following lemma shows that $\Theta_k^{(0)}$ is indeed a good initialization for $\Theta_k^*$. 

\begin{lemma}
\label{lemma:joint_initialization}
Suppose the condition (OC) is satisfied, then the initialization $\Theta_k^{(0)}$ satisfies
\begin{equation}
\big\| \Theta_k^{(0)} - \Theta_k^* \big\|_F \leq 2\tilde CK \rho_0 + (\eta-1)\sigma_{\max},
\end{equation}
for some constant $\tilde C$ where $\sigma_{\max} = \max_k \|\Theta_k^*\|_2$. 
\end{lemma}

The initialization $\Theta_k^{(0)}$ is no longer $\sqrt n$-consistent. Nevertheless it is not required. 
With this initialization, we then follow Algorithm \ref{algo:joint} and estimate $B_1,B_2$ and $M$ alternatively. Note that when estimating $B_1$ and $B_2$, we run Algorithm \ref{algo:B1B2} for large enough $T$ so that the first term in \eqref{eq:linear_convergence} is small compared to the second term. These $T$ iterations for Algorithm \ref{algo:B1B2} are one iteration for Algorithm \ref{algo:joint} and we use $B^{[t]} = [B_1^{[t]}, B_2^{[t]}]$ and $M^{[t]}$ to denote the iterates we obtained from Algorithm \ref{algo:joint}. 
Denote $d^2(M, M^*) = \frac{1}{n} \sum_{i=1}^n \sum_{k_0=1}^{K} (m_{ik_0} - m_{ik_0}^*)^2$. We obtain the following theorem on estimation error for jointly learning.

%
\begin{theorem}
\label{theorem:joint}
Suppose the conditions in Theorem \ref{theorem:known} hold and suppose condition (DC) and (OC) hold. For large enough $n$, after $T$ iterations of Algorithm \ref{algo:joint} we have 
\begin{equation}
\begin{aligned}
d^2\big(B^{[T]},B^*\big) &+ d^2\big(M^{[T]}, M^*\big) \leq \frac{ C_1 e_{{\rm stat}, M}^2 + C_2 e_{{\rm stat}, \Theta}^2 }{1-\beta_0}  + \beta_0^T \Big[ d^2\big(B^{[0]},B^*\big) + d^2\big(M^{[0]}, M^*\big) \Big],
\end{aligned}
\end{equation}
for some constant $\beta_0 < 1$, which shows linear convergence up to statistical error.
\end{theorem}

%
%

%% file: Synthetic.tex
\section{Simulation}
\label{sec:synthetic}

In this section we evaluate our model and algorithms on synthetic
datasets. We first consider the setting where the topics are known and
we consider $p=200$ nodes with $K=10$ topics. The true matrices
$B_1^*$ and $B_2^*$ are generated row by row where we randomly select
1-3 topics for each row and set a random value generated from
${\sf Uniform}(1, 2)$. All the other values are set to be 0.  This
gives sparsity level $s^* = 2p = 400$ in expectation, and we set
$s = 2s^*$ in the algorithm as the hard thresholding parameter.  For
each observation, we randomly select 1-3 topics and assign each
selected topic a random value ${\sf Uniform}(0, 1)$, and 0
otherwise. We then normalize this vector to get the topic distribution
$m_i$.  The true value $X_i^*$ is generated according to
\eqref{eq:Xistar}. Note that $X_i^*$ is also a sparse matrix. We
consider two types of observation: real valued observation and binary
valued observation. For real valued observation, we generate $X_i$
(equivalently, set $E_i$) in the following way: first we randomly
select 10\% of the nonzero values in $X_i^*$ and set to 0 (miss some
edges); second for each of the remaining nonzero values, we generate
an independent random number ${\sf Uniform}(0.3, 3)$ and multiply with
the original value (observe edges with noise); finally we randomly
select 10\% of the zero values in $X_i^*$ and set them as
${\sf Uniform}(0, 1)$ (false positive edges).  For binary
observations, we treat the true values in $X_i^*$ as probability of
observing an edge, and generate $X_i$ as
$X_i = {\sf Bernoulli}(X_i^*)$.
For those true values greater than 1 we just set $X_i$ to be
1. Finally we again pick 10\% false positive edges.

\begin{figure*}[htbp]
\begin{center}
\begin{minipage}[t]{0.48\linewidth}
\centering
\includegraphics[width=0.95\textwidth]{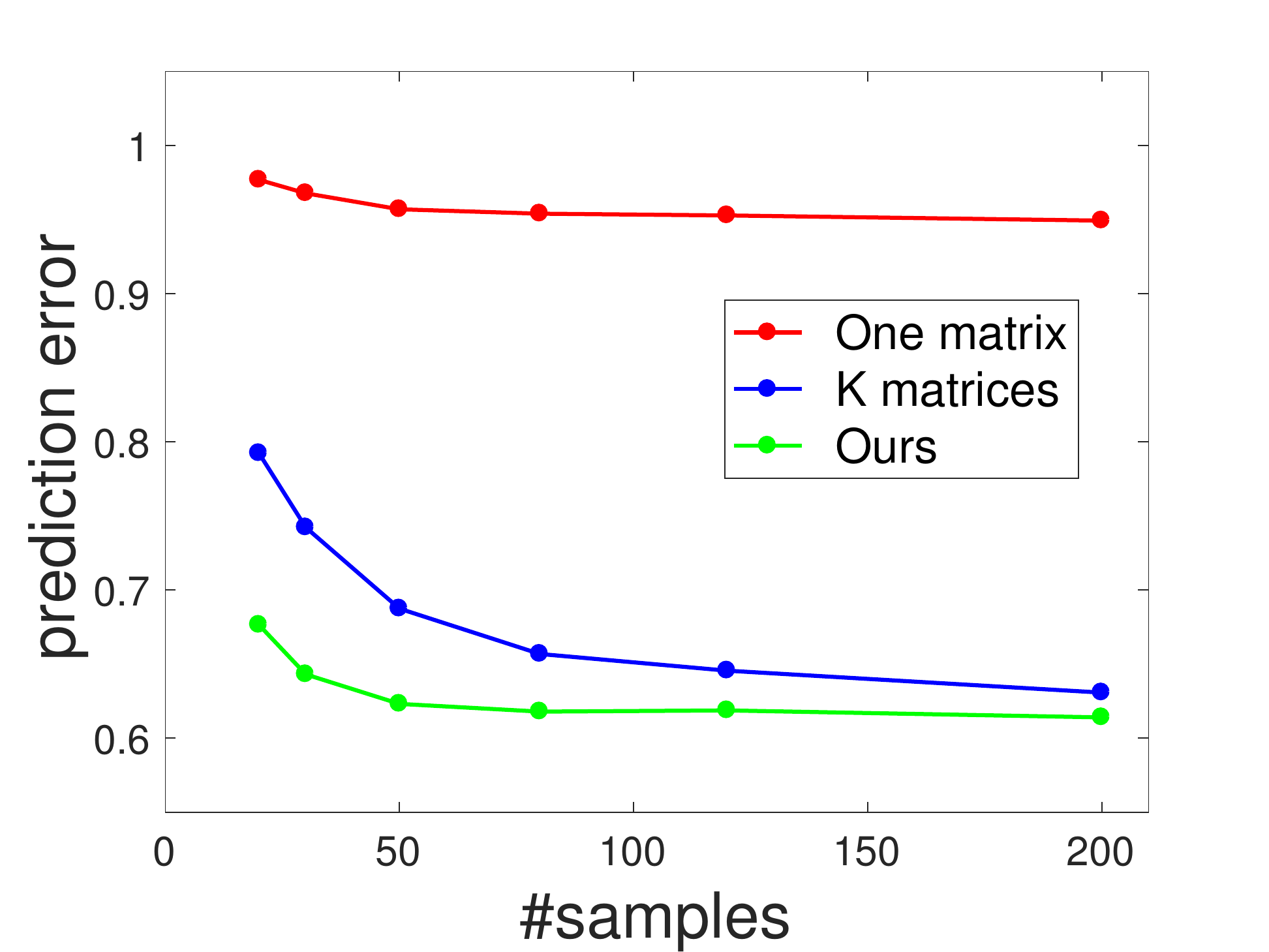}
\caption{Prediction error for real-valued observation, with known topics}
\label{fig:know_real}
\end{minipage}
\begin{minipage}[t]{0.48\linewidth}
\centering
\includegraphics[width=0.95\textwidth]{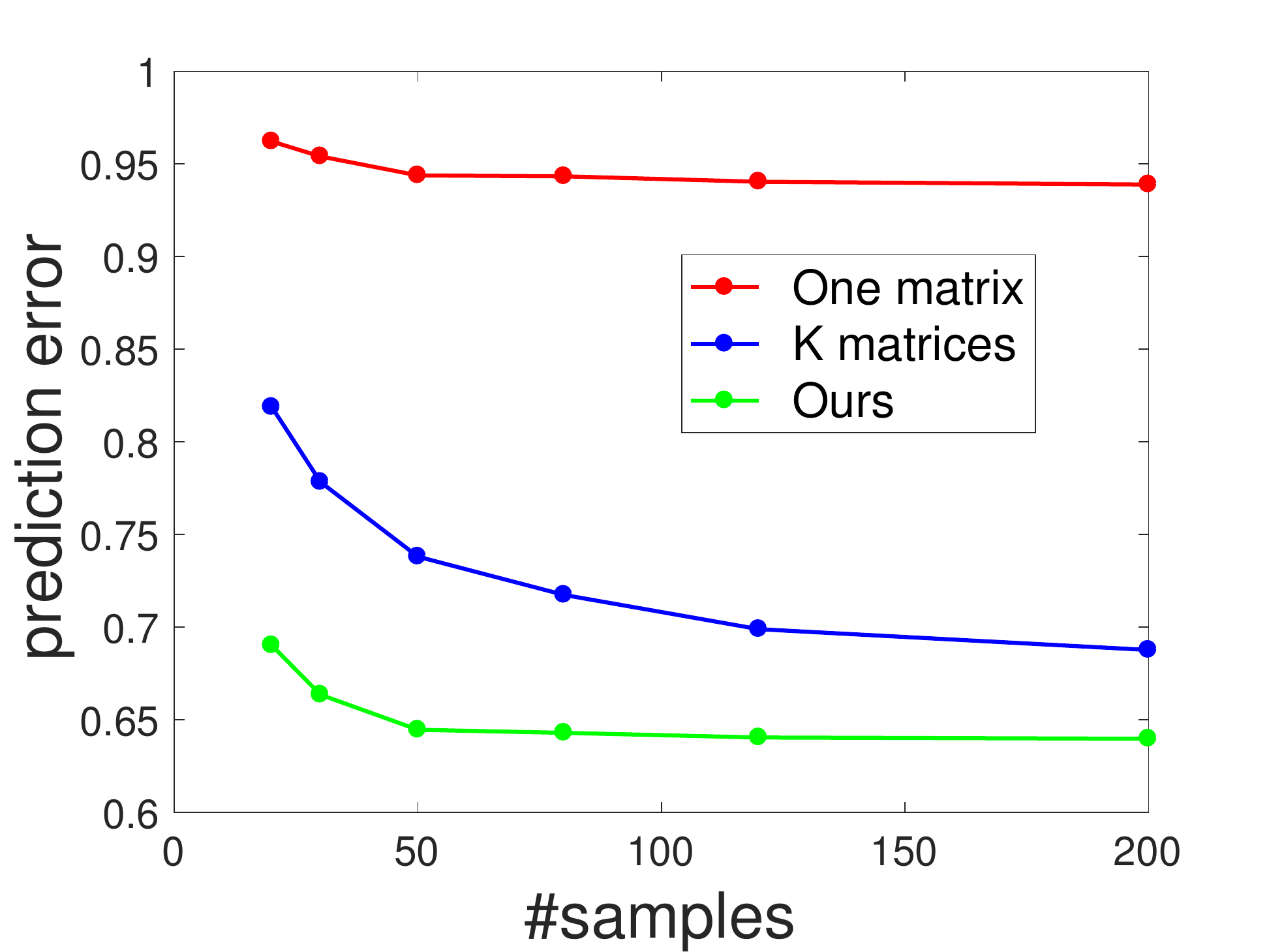}
\caption{Prediction error for binary observation, with known topics}
\label{fig:know_binary}
\end{minipage}
\end{center}
\end{figure*}

We vary the number of observations
$n \in \{20, 30, 50, 80, 120, 200\}$ and compare our model with the
following two state-of-the-art methods. The first method is inspired by 
\cite{gomez2010inferring} which ignores the topic information and
uses one $p \times p$ matrix to capture the entire dataset (termed
``One matrix''). This matrix is given by $\overline X$. The second
method is inspired by \cite{du2013uncover} which considers the topic
information and assigns each topic a $p \times p$ matrix (termed ``$K$
matrices''). However it still ignores the node-topic structure. For
this model, we ignore the rank constraint and return the matrix
$\Theta_k$ given by the initialization procedure.  Note that ``One
matrix'' method has $p^2$ parameters, ``$K$ matrices'' has $p^2K$
parameters, but our method has only $2pK$ parameters. Since we usually
have $K \ll p$, we are able to use much fewer parameters
to capture the network structure, and would not suffer too much from
overfitting.  
For fair comparison, we also do hard thresholding on
each of these $p \times p$ matrices with parameter $4p$.  
The comparison is done by evaluating the objective function on independent
test dataset (prediction error). 
This prediction error is given by $\frac{1}{n} \sum_i ||X_i - \hat X_i||_F^2$, where $X_i$ is the observed value and $\hat X_i$ is the predicted value. The predicted values take different forms for each method. For ``One matrix'' it is just $\overline X$; for ``$K$ matrices'' it is the weighted sum of the $K$ estimated matrices for each topic; for our model, the prediction is obtained by plugging in the estimated $B_1$ and $B_2$ into \eqref{eq:Xistar}.
Figure \ref{fig:know_real} and Figure
\ref{fig:know_binary} show
the comparison results for real valued and binary
observation, respectively. Each result is based on 20 replicates.  We
can see that our method has the best prediction error since we are
able to utilize the topic information and the structure among nodes
and topics; ``One matrix'' method completely ignores the topic
information and ends up with bad prediction error; $K$ matrices''
method ignores the structure among nodes and topics and suffers from
overfitting. As sample size goes large, ``$K$ matrices'' method will
behave closer to our model in terms of prediction error, since our
model is a special case of the $K$ matrices model. However, it still
cannot identify the structure among nodes and topics and is hard to
interpret.

\begin{figure*}[htbp]
\begin{center}
\begin{minipage}[t]{0.48\linewidth}
\centering
\includegraphics[width=0.95\textwidth]{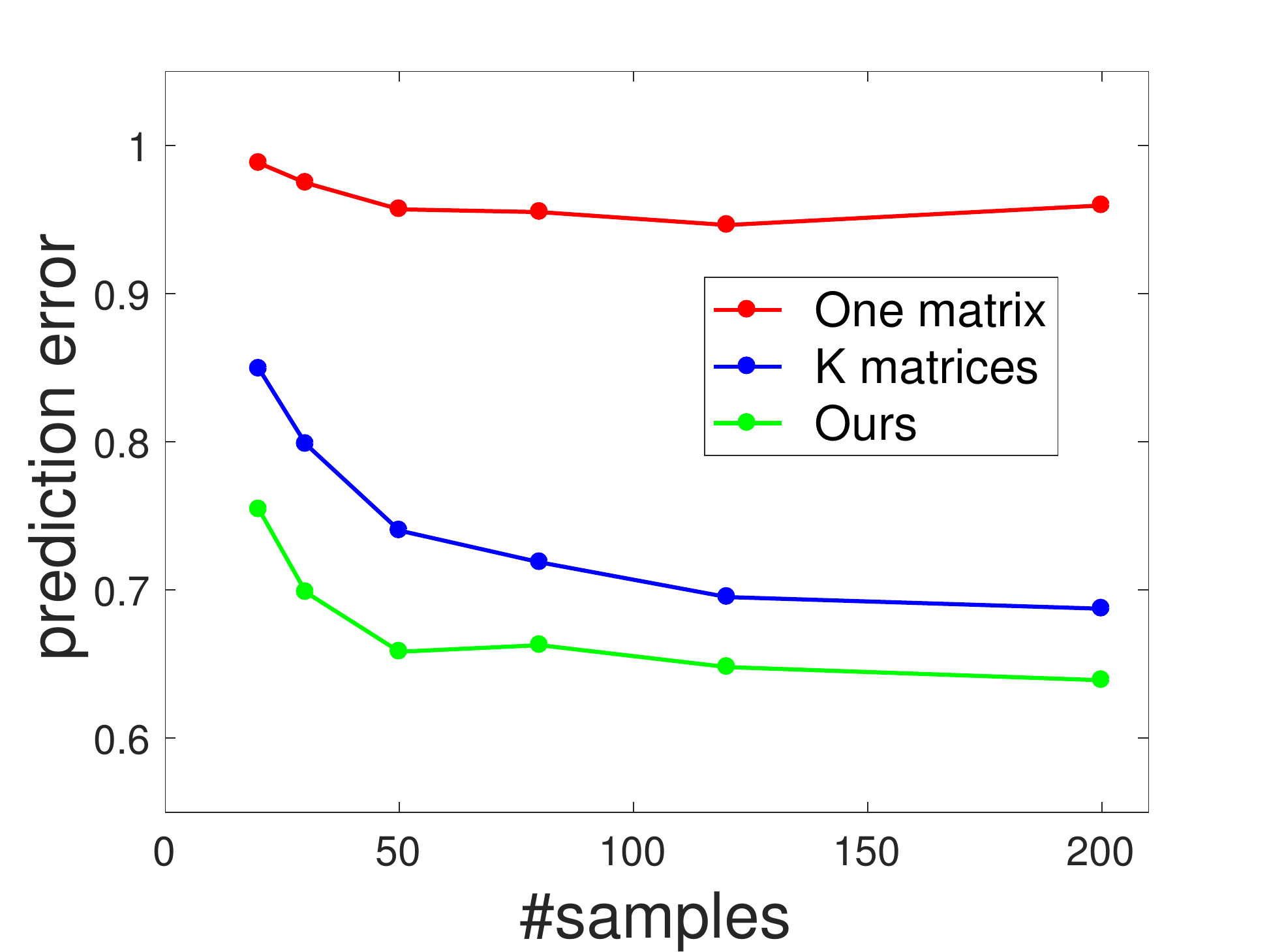}
\caption{Prediction error for real-valued observation, with unknown topics}
\label{fig:unknown_real}
\end{minipage}
\begin{minipage}[t]{0.48\linewidth}
\centering
\includegraphics[width=0.95\textwidth]{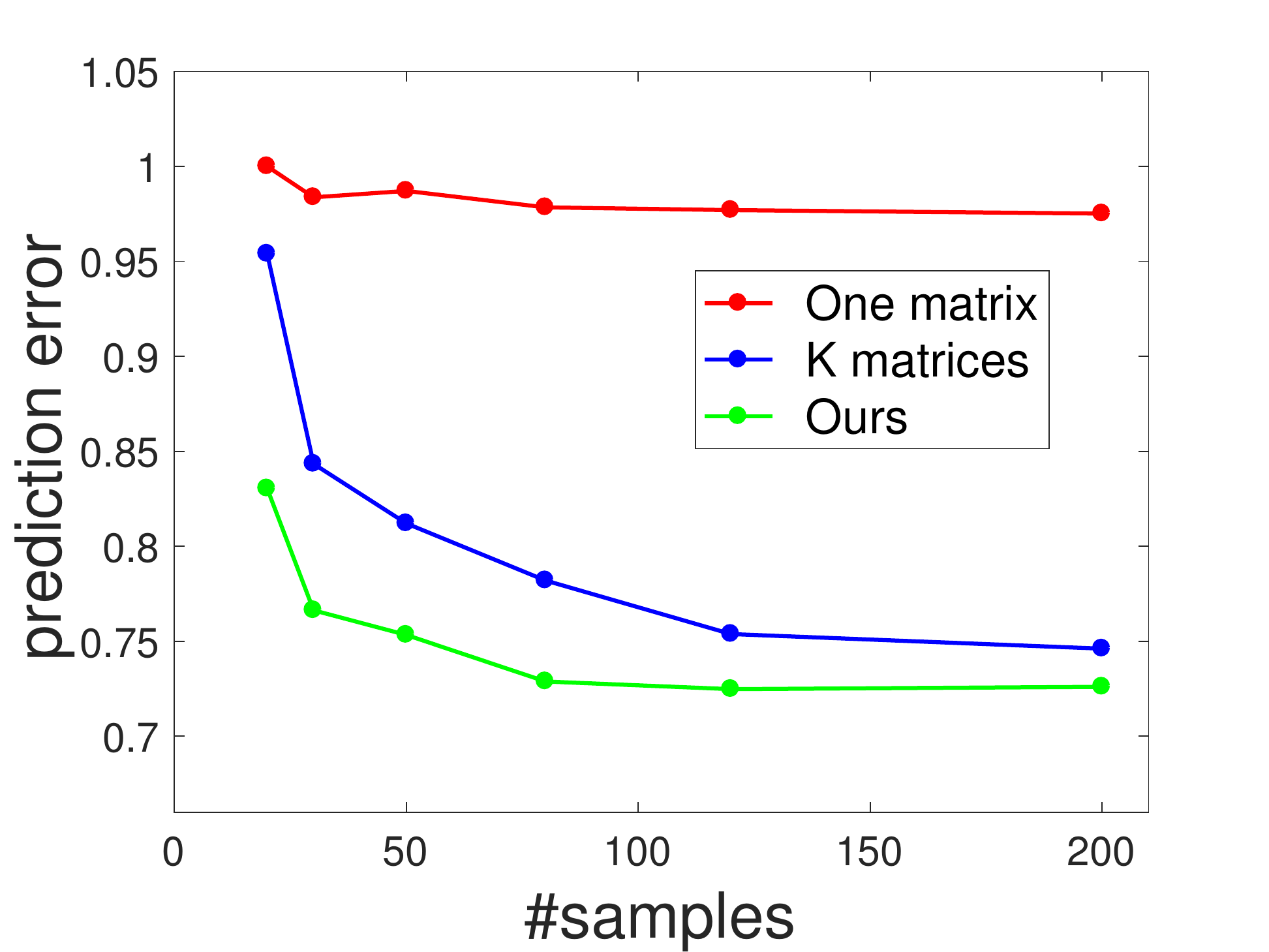}
\caption{Prediction error for binary observation, with unknown topics}
\label{fig:unknown_binary}
\end{minipage}

\end{center}
\end{figure*}

We then consider the setting where the topics are unknown. We
initialize and estimate $B_1,B_2$ and $M$ according to the procedure
described in Section \ref{sec:joint}; for ``One matrix'' method, the
estimator is still given by $\overline X$; for ``$K$ matrices''
method, we estimate $\Theta$ and $M$ by alternating gradient method on
the objective function \eqref{eq:loss_Theta_M}. All the other setups
are the same as the previous case. Figure \ref{fig:unknown_real} and
Figure \ref{fig:unknown_binary} show the comparison results for real
valued and binary observation, respectively.  Again we
see that our model behaves the best. 
These results
demonstrate the superior performance of our model and algorithm
compared with existing state-of-the-art methods.

Finally we check the running time of our method experimentally. Here we fix $n=500, T=50$ and vary $K$ and $p$. 
The empirical running time is given in Table \ref{running_time}, where we see a linear dependency on $K$ and quadratic dependency on $p$, in line with the claim in remark \ref{remark:complexity}.

\begin{table}
\caption{Running time (in second) with different $K$ and $p$}
\label{running_time}
\begin{center}
\begin{small}
\begin{tabular}{lccccc}
\hline
\\[-2.3ex]
 & $K=10$  & $K=20$ & $K=40$ \\\hline
 \\[-1.9ex]
$p=100$ & 1.7 & 2.2 & 3.1 \\
$p=200$ & 4.1 & 5.2 & 7.5 \\
$p=400$ & 13.0 & 16.0 & 22.0 \\
\\[-2.3ex]
\hline
\end{tabular}
\end{small}
\end{center}
\end{table}

%% file: Real.tex
\section{Application to ArXiv data}
\label{sec:real}

In this section we evaluate our model on real dataset. The dataset we use is the ArXiv collaboration and citation network dataset on high energy physics theory \cite{leskovec2005graphs, gehrke2003overview}. This dataset covers papers uploaded to ArXiv high energy physics theory category in the period from 1993 to 2003, and the citation network for each paper. For our experiment we treat each author as a node and each publication as an observation. 
For each publication $i$, we set the observation matrix $X_i$ in the following way: the component $(x_i)_{j\ell} = 1$ if this paper is written by author $j$ and cited by author $\ell$, and $(x_i)_{j\ell} = 0$ otherwise. Since each paper has only a few authors, we consider a variant of our original model as
\begin{equation}
X_i^* = \big[ B_1  M_i  B_2^\top \big] \, \odot \, A_i
\label{eq:Xistar_citation}
\end{equation}
where operator $\odot$ is component-wise product and $A_i \in \RR^{p \times p}$ is an indictor matrix with $(a_i)_{j\ell} = 1$ if $j$ is the author of this paper, and $(a_i)_{j\ell} = 0$ otherwise. This means for each paper, we only consider the influence behavior of its authors. 

For our experiment we consider the top 200 authors with about top 10000 papers in terms of number of citations, and split the papers into 8000 training set and 2000 test set. 
We first do Topic modeling on the abstracts of all the papers and extract $K=6$ topics as well as the topic distribution on each paper. We then treat this topic information as \emph{known} and apply our Algorithm \ref{algo:B1B2} to the training set and learn the two node-topic matrices. 
These two matrices are given in Table \ref{B1_citation_detail} and Table \ref{B2_citation_detail}. 
The keywords of the 6 topics are shown at the head of the two tables and the first column of the two tables is the name
of the author.

We then compare the node-topic structure to the research interests and publications listed by the authors themselves on their website. The comparison results show that our model is able to capture the research topics accurately. 
For example, 
Christopher Pope reports quantum gravity and string theory; 
Arkady Tseytlin reports quantum field theory; 
Emilio Elizalde reports quantum physics;
Cumrun Vafa reports string theory;
Ashoke Sen reports string theory and black holes as their research areas in their webpages. 
These are all successfully captured by our method.

Finally we compare the result with ``One matrix'' and ``$K$ matrices'' methods on test set. The comparison result is given in Table \ref{compare_citation} for training error, testing error, number of total parameters, and number of nonzero parameters. 
Since our model has much fewer parameters, it has the largest training error. However we can see that our model has the best test error, and both the other two methods do not generalize to test set and suffer from overfitting. 
These results demonstrates that the topic information and node-topic structure do exist, and our model is able to capture them.

\begin{table*}
\caption{ The influence matrix $B_1$ for citation dataset}
\vspace{-2mm}
\label{B1_citation_detail}
\begin{center}
\begin{small}
\setlength{\tabcolsep}{3.8pt}
\begin{tabular}{lcccccc}
\hline
\\[-2ex]
 & \begin{tabular}{@{}c@{}} black \\ hole \\ energy \\ chains \end{tabular} 
 & \begin{tabular}{@{}c@{}} quantum \\ model \\ field \\ theory \end{tabular} 
 & \begin{tabular}{@{}c@{}} gauge \\ theory \\ field \\ effective \end{tabular} 
 & \begin{tabular}{@{}c@{}} algebra \\ space \\ group \\ structure \end{tabular} 
 & \begin{tabular}{@{}c@{}} states \\ space \\ noncommutative \\ boundary \end{tabular} 
 & \begin{tabular}{@{}c@{}} string \\ theory \\ supergravity \\supersymmetric \end{tabular} \\
 \\[-1.9ex]
\hline
\\[-1.9ex]
Christopher Pope&       &   0.359&       & 0.468&  &   0.318\\
Arkady Tseytlin&   0.223&   0.565&  &  0.25&       &       \\
Emilio Elizalde&       &   0.109&       &     &  & \\
Cumrun Vafa&       &       &    0.85& 0.623&   0.679&   0.513\\
Edward Witten&       &   0.204&       & 0.795&   0.678&    1.87\\
Ashok Das& &   0.155&   0.115&  1.07&       &       \\
Sergei Odintsov&       &    &       &     &  &   \\
Sergio Ferrara&   0.297&   0.889&   0.345& 0.457&   0.453&   0.249\\
Renata Kallosh&    0.44&   0.512&       & 0.326&   0.382&       \\
Mirjam Cvetic&       &   0.339&   0.173& 0.338&       &       \\
Burt A. Ovrut&   0.265&   0.191&   0.127& 0.328&   0.133&       \\
Ergin Sezgin&  &    0.35&       & 0.286&       &       \\
Ian I. Kogan&  &    &   0.193&     &  &  \\
Gregory Moore&       &   0.323&    0.91& 0.325&   0.536&       \\
I. Antoniadis&   0.443&   0.485&  & 0.545&   0.898&   0.342\\
Mirjam Cvetic&   0.152&   0.691&  & 0.228&   0.187&       \\
Andrew Strominger&   0.207&   0.374&   0.467&  1.15&       &       \\
Barton Zwiebach&    0.16& &  & 0.222&   0.383&   0.236\\
P.K. Townsend&       &   0.629&  & 0.349&       &  0.1\\
Robert C. Myers&       &   0.439& &  0.28& &  \\
E. Bergshoeff&  &   0.357&       & 0.371&       &  \\
Amihay Hanany&  &   0.193&       & 0.327&       &    1.09\\
Ashoke Sen&   0.319&   &    & 0.523&       &   0.571\\
\\[-2ex]
\hline
\end{tabular}
\end{small}
\end{center}
\vskip -0.1in
\end{table*}

\begin{table*}
\caption{ The receptivity matrix $B_2$ for citation dataset}
\vspace{-2mm}
\label{B2_citation_detail}
\begin{center}
\begin{small}
\setlength{\tabcolsep}{3.8pt}
\begin{tabular}{lcccccc}
\hline
\\[-2ex]
 & \begin{tabular}{@{}c@{}} black \\ hole \\ energy \\ chains \end{tabular} 
 & \begin{tabular}{@{}c@{}} quantum \\ model \\ field \\ theory \end{tabular} 
 & \begin{tabular}{@{}c@{}} gauge \\ theory \\ field \\ effective \end{tabular} 
 & \begin{tabular}{@{}c@{}} algebra \\ space \\ group \\ structure \end{tabular} 
 & \begin{tabular}{@{}c@{}} states \\ space \\ noncommutative \\ boundary \end{tabular} 
 & \begin{tabular}{@{}c@{}} string \\ theory \\ supergravity \\supersymmetric \end{tabular} \\
 \\[-1.9ex]
\hline
\\[-1.9ex]
Christopher Pope& 0.477& 0.794&     &  0.59&     &     \\
Arkady Tseytlin& 0.704&  1.16& 0.312& 0.487&     & 0.119\\
Emilio Elizalde&     &     &     &     &     &     \\
Cumrun Vafa& 0.309&     & 0.428& 0.844& 0.203& 0.693\\
Edward Witten& 0.352&     & 0.554& 0.585& 0.213& 0.567\\
Ashok Das& 0.494& 0.339&     & 0.172&     &     \\
Sergei Odintsov&     & 0.472&     &     &     &     \\
Sergio Ferrara& 0.423&  0.59& 0.664& 0.776&     &     \\
Renata Kallosh& 0.123& 0.625& 0.638& 0.484&     & 0.347\\
Mirjam Cvetic&  0.47& 0.731&     & 0.309&     &     \\
Burt A. Ovrut& 0.314& 0.217&  0.72& 0.409&     & 0.137\\
Ergin Sezgin&     & 0.108& 0.161& 0.358&     &     \\
Ian I. Kogan& 0.357& 0.382&     &     &     & 0.546\\
Gregory Moore& 0.375& 0.178& 0.721&  0.69& 0.455& 0.517\\
I. Antoniadis& 0.461&     & 0.699& 0.532&     & 0.189\\
Mirjam Cvetic& 0.409&  1.11& 0.173& 0.361&     &     \\
Andrew Strominger&     & 0.718& 0.248& 0.196& 0.133&     \\
Barton Zwiebach&     &     & 0.308& 0.204&     & 0.356\\
P.K. Townsend& 0.337& 0.225& 0.245& 0.522&     &     \\
Robert C. Myers& 0.364& 0.956&     & 0.545&     & 0.139\\
E. Bergshoeff& 0.487& 0.459& 0.174& 0.619&     &     \\
Amihay Hanany& 0.282&     & 0.237& 0.575&     & 0.732\\
Ashoke Sen&     & 0.214&  0.18&  0.37&     &     \\
\\[-2ex]
\hline
\end{tabular}
\end{small}
\end{center}
\vskip -0.1in
\end{table*}

\begin{table}
\caption{Comparison of the 3 methods on test links for citation dataset}
\vspace{-2mm}
\label{compare_citation}
\begin{center}
\begin{small}
\begin{tabular}{lcccccc}
\hline
\\[-2.3ex]
 & train  & test & $\#$ para &  $\#$ nonzero \\\hline
 \\[-1.9ex]
One matrix \cite{gomez2010inferring} & 7.628 & 8.223 & 40000 & 7695 \\
$K$ matrices \cite{du2013uncover} & 5.861 & 8.415 &  240000 &  19431 \\
Our method & 8.259 & 8.217 & 2400  & 1200  \\
\\[-2.3ex]
\hline
\end{tabular}
\end{small}
\end{center}
\end{table}

%% file: Conclusion.tex
\section{Conclusion}
\label{sec:conclusion}

In this paper we propose an influence-receptivity model and show how
this structure can be estimated with theoretical
guarantee. Experiments show superior performance of our model on 
synthetic and real data, compared with existing methods. This
influence-receptivity model also provides much better
interpretability.

There are several future directions we would like to
pursue.
Currently the topic information is either learned from
topic modeling and fixed, or is (jointly) learned by our model where
we ignore the text information.  It would be of interest to combine
the influence-receptivity structure and topic modeling to provide more
accurate results.  Another extension would be allowing dynamic
influence-receptivity structure over time.  

%% file: Appendix.tex
\section{Technical proofs}

\subsection{Proof of Lemma \ref{lemma:init}.}

\begin{proof}
Since $\hat\Theta = (\hat \Theta_1, ..., \hat \Theta_K)$ are the global minimum of \eqref{eq:loss_Theta}, we have
\begin{equation}
0 \geq f(\hat\Theta) - f(\Theta^*) \geq \big\langle \nabla f(\Theta^*), \hat\Theta - \Theta^* \big\rangle + \frac{\mu_\Theta}{2} \big\| \hat\Theta - \Theta^* \big\|_F^2.
\end{equation}
We then have
\begin{equation}
\big\| \hat\Theta - \Theta^* \big\|_F^2 \leq -\frac{2}{\mu_\Theta} \big\langle \nabla f(\Theta^*), \hat\Theta - \Theta^* \big\rangle \leq \frac{2}{\mu_\Theta} \big\| \nabla f(\Theta^*) \big\|_F \cdot \big\| \hat\Theta - \Theta^* \big\|_F,
\end{equation}
and hence
\begin{equation}
\big\| \hat\Theta - \Theta^* \big\|_F \leq \frac{2}{\mu_\Theta} \big\| \nabla f(\Theta^*) \big\|_F.
\end{equation}

\end{proof}

\subsection{Proof of Theorem \ref{theorem:known}.}

\begin{proof}
We apply the non-convex optimization result in \cite{yu2018recovery}. Since the initialization condition and (RSC/RSS) are satisfied for our problem according to Lemma \ref{lemma:init}, we apply Lemma 3 in \cite{yu2018recovery} and obtain
\begin{equation}
\label{eq:lemma_contraction}
d^2 \Big( B^{(t+1)} , B^* \Big) \leq \xi^2 \bigg[ \Big(1 - \eta\cdot
\frac{2}{5} \mu_{\min}\sigma_{\max}
\Big)\cdot d^2 \Big( B^{(t)}, B^* \Big)
 +
\eta\cdot\frac{L_{\Theta}  + \mu_{\Theta} }{L_{\Theta} \cdot\mu_{\Theta} }\cdot e_{{\rm stat},\Theta}^2 \bigg],
\end{equation}
where $\xi^2 = 1 + \frac{2}{\sqrt{c - 1}}$ and $\sigma_{\max} = \max_k \|\Theta_k^*\|_2$. Define the contraction value
\begin{equation}
\label{eq:def_beta}
\beta = \xi^2 \Big(1 - \eta\cdot
\frac{2}{5} \mu_{\min}\sigma_{\max} \Big) < 1, 
\end{equation}
we can iteratively apply \eqref{eq:lemma_contraction} for each $t = 1, 2, ..., T$ and obtain
\begin{equation}
d^2 \Big( B^{(T)}, B^* \Big) \leq \beta^T d^2 \Big( B^{(0)}, B^* \Big) +  \frac{\xi^2\eta}{1-\beta}\cdot\frac{L_{\Theta}  + \mu_{\Theta} }{L_{\Theta} \cdot\mu_{\Theta}}\cdot e_{{\rm stat},\Theta}^2,
\end{equation}
which shows linear convergence up to statistical error.
\end{proof}

\subsection{Proof of Lemma \ref{lemma:joint_initialization}.}

\begin{proof}
Since $\tilde X$ is the best rank $K$ approximation for $\overline X$ and $\overline X^*$ is also rank $K$, we have $\| \tilde X - \overline X \|_F \leq \| \overline X^* - \overline X \|_F $ and hence
\begin{equation}
\label{eq:2error}
\| \tilde X - \overline X^* \|_F \leq \| \tilde X - \overline X \|_F + \| \overline X^* - \overline X \|_F \leq 2\| \overline X^* - \overline X \|_F = 2\|\overline E\|_F.
\end{equation}
By definition we have
\begin{equation}
 \overline X^* = \sum_{k=1}^K \Big(\frac{1}{n} \sum_{i=1}^n m_{ik}^* \Big) \Theta_k^* = B_1^* A^* {B_2^*}^\top = Q_1R_1A^*R_2^\top Q_2^\top = Q_1 (A_{\rm diag} + A_{\rm off}) Q_2^\top.
\end{equation}
Plug back to \eqref{eq:2error} we obtain
\begin{equation}
\big\| \tilde X - Q_1 (A_{\rm diag} + A_{\rm off}) Q_2^\top \big\|_F \leq 2\|\overline E\|_F, 
\end{equation}
and hence
\begin{equation}
\label{eq:svd_diff}
\Big\| \sum_{k=1}^K \tilde \sigma_k \tilde u_k \tilde v_k^\top - Q_1 A_{\rm diag} Q_2^\top \Big\|_F \leq 2\|\overline E\|_F + \|Q_1 A_{\rm off} Q_2^\top\|_F \leq 2\|\overline E\|_F + \rho_0.
\end{equation}
Since $\overline E$ is the mean value of i.i.d. errors $E_i$, we have that $\|\overline E\|_F \propto n^{-1/2}$ and therefore can be arbitrarily small with large enough $n$. 
Moreover, the left hand side of \eqref{eq:svd_diff} is the difference of two singular value decompositions. According to the matrix perturbation theory, for each $k$ we have (up to permutation)
\begin{equation}
\Big\| \tilde \sigma_k \tilde u_k \tilde v_k^\top - q_{1,k} \cdot a_{{\rm diag},k} \cdot q_{2,k}^\top \Big\|_F \leq 2C \rho_0,
\end{equation}
and hence
\begin{equation}
\Big\| \tilde \sigma_k \tilde u_k \tilde v_k^\top - \frac{1}{n} \sum_{i=1}^n m_{ik}^* \Theta_k^* \Big\|_F \leq 2 \tilde C \rho_0.
\end{equation}
Finally we obtain
\begin{equation}
\Big\| K \cdot \tilde \sigma_k \tilde u_k \tilde v_k^\top - \Theta_k^* \Big\|_F = K \cdot \Big\| \tilde \sigma_k \tilde u_k \tilde v_k^\top - \frac{1}{K}\Theta_k^* \Big\|_F \leq K \cdot \Big(2 \tilde C \rho_0 + \Big|\frac{1}{n} \sum_{i=1}^n m_{ik}^* - \frac{1}{K} \Big| \cdot \|\Theta_k^*\|_F \Big)
\leq 2\tilde CK \rho_0 + (\eta-1)\sigma_{\max}.
\end{equation}
\end{proof}

\subsection{Proof of Theorem \ref{theorem:joint}}
We analyze the two estimation step in Algorithm \ref{algo:joint}.
\paragraph{Update on $B_1$ and $B_2$.} 
The update algorithm on $B_1$ and $B_2$ is the same with known $M$. Besides the statistical error defined in \eqref{eq:def_stat_error}, we now have an additional error term due to the error in $M$. 
Recall that $d^2(M, M^*) = \frac{1}{n} \sum_{i=1}^n \sum_{k_0=1}^{K} (m_{ik_0} - m_{ik_0}^*)^2$, 
Lemma \ref{lemma:joint_Theta} quantifies the effect of one estimation step on $B$.

\begin{lemma}
\label{lemma:joint_Theta}
Suppose the conditions in Theorem \ref{theorem:known} hold and suppose condition (DC) and (OC) hold, we have
\begin{equation}
d^2 \big( B^{[t]}, B^* \big) \leq C_1 \cdot e^2_{\rm{stat}, \Theta} + \beta_1 \cdot d^2\big(M^{[t]}, M^*\big),
\end{equation}
for some constant $C_1$ and $\beta_1$.
\end{lemma}

\paragraph{Update on $M$.} Lemma \ref{lemma:joint_M} quantifies the effect of one estimation step on $M$.
\begin{lemma}
\label{lemma:joint_M}
Suppose the condition (TC) holds, we have
\begin{equation}
d^2\big(M^{[t]}, M^*\big) \leq C_2 \cdot e_{{\rm{stat}}, M}^2 + \beta_2 \cdot d^2 \big( B^{[t]}, B^* \big),
\end{equation}
for some constant $C_2$ and $\beta_2$.
\end{lemma}

Denote $\beta_0 = \min\{\beta_1, \beta_2\}$, as long as the signal $\sigma_{\max}$ is small and the noise $E_i$ is small enough we can guarantee that $\beta_0 < 1$. 
Combine Lemma \ref{lemma:joint_Theta} and \ref{lemma:joint_M} we complete the proof.

\subsection{Proof of Lemma \ref{lemma:joint_Theta}.}

\begin{proof}
The analysis is exactly the same with the case where $M$ is known except that the statistical error is different. Specifically, for each $k$ we have
\begin{equation}
\begin{aligned}
\nabla_{\Theta_k} f(\Theta^*, M) &= -\frac{1}{n} \sum_{i=1}^n \Big(X_i - \sum_{k_0=1}^{K} m_{ik_0} \Theta_{k_0}^* \Big) \cdot m_{ik} \\
&= -\frac{1}{n} \sum_{i=1}^n \Big(E_i + \sum_{k_0=1}^{K} (m_{ik_0}^* - m_{ik_0}) \Theta_{k_0}^* \Big) \cdot m_{ik} \\
&= \underbrace{ -\frac{1}{n} \sum_{i=1}^n E_i m_{ik}^*}_{R_1} +
 \underbrace{ \frac{1}{n} \sum_{i=1}^n E_i (m_{ik}^* - m_{ik}) }_{R_2}+ 
 \underbrace{ \frac{1}{n} \sum_{i=1}^n \sum_{k_0=1}^{K} (m_{ik_0} - m_{ik_0}^*) \Theta_{k_0}^* \cdot m_{ik} }_{R_3}.
\end{aligned}
\end{equation}
The first term $R_1$ is just the usual statistical error term on $\Theta$. For term $R_2$, denote $e_0 = \frac{1}{n} \sum_{i=1}^n \|E_i\|_F^2$, 
we have
\begin{equation}
\|R_2\|_F^2 \leq \frac{1}{n^2} \Big( \sum_{i=1}^n \|E_i\|_F^2 \Big) \cdot \sum_{i=1}^{n} (m_{ik} - m_{ik}^*)^2 \leq \frac{e_0}{n} \sum_{i=1}^{n} (m_{ik} - m_{ik}^*)^2.
\end{equation}
For term $R_3$, we have
\begin{equation}
\begin{aligned}
\|R_3\|_F^2 &\leq \frac{1}{n^2} \Big\| \sum_{i=1}^n \sum_{k_0=1}^{K} (m_{ik_0} - m_{ik_0}^*) \Theta_{k_0}^* \cdot m_{ik} \Big\|_F^2
\leq \frac{1}{n^2} \bigg( \sum_{i=1}^n \sum_{k_0=1}^{K} (m_{ik_0} - m_{ik_0}^*)^2 \bigg) \cdot \bigg(  \sum_{i=1}^n \sum_{k_0=1}^K \| \Theta_{k_0}^*  \|_F^2 \cdot m_{ik} ^2 \bigg) \\
&\leq \frac{K \sigma_{\max}^2}{n^2} \bigg( \sum_{i=1}^nm_{ik}^2 \bigg) \cdot  \bigg( \sum_{i=1}^n \sum_{k_0=1}^{K} (m_{ik_0} - m_{ik_0}^*)^2 \bigg).
\end{aligned}
\end{equation}
Taking summation over all $k$, the first term $R_1$ gives the statistical error as before, the terms $R_2$ and $R_3$ gives
\begin{equation}
\begin{aligned}
\sum_{k=1}^K \|R_2\|_F^2 + \|R_3\|_F^2 \leq \frac{e_0+K\sigma_{\max}^2}{n}
\bigg( \sum_{i=1}^n \sum_{k=1}^{K} (m_{ik} - m_{ik}^*)^2 \bigg).
\end{aligned}
\end{equation}
\end{proof}

\subsection{Proof of Lemma \ref{lemma:joint_M}.}

\begin{proof}
The estimation on $M$ is separable with each $m_i$. Denote the objective function on observation $i$ as 
\begin{equation}
\label{eq:loss_i_Theta_M}
f_i(\Theta, m_i) =  \Big\|X_i - \sum_{k=1}^K m_{ik} \cdot \Theta_k \Big\|_F^2.
\end{equation}
According to condition (DC), the objective function \eqref{eq:loss_i_Theta_M} is $\mu_M$-strongly convex in $m_i$. 
Similar to the proof of Lemma \ref{lemma:init}, we obtain
\begin{equation}
\sum_{k=1}^K (m_{ik} - m_{ik}^*)^2 \leq \frac{4}{\mu_M^2} \big\| \nabla_{m_i} f_i(\Theta, m_i^*) \big\|_F^2
= \frac{4}{\mu_M^2} \sum_{k=1}^K \Big[ \nabla_{m_{ik}} f_i(\Theta, m_i^*) \Big]^2.
\end{equation}
Moreover, we have
\begin{equation}
\begin{aligned}
\nabla_{m_{ik}} f_i(\Theta, m_i^*) &= - \Big\langle X_i - \sum_{k_0=1}^K m_{ik_0}^* \cdot \Theta_{k_0}, \Theta_k \Big\rangle \\
&= \underbrace{ - \Big\langle E_i, \Theta_k^* \Big\rangle }_{T_1}
+ \underbrace{ \Big\langle E_i, (\Theta_k^* - \Theta_k) \Big\rangle }_{T_2}
+ \underbrace{ \Big\langle \sum_{k_0=1}^K m_{ik_0}^* (\Theta_{k_0} - \Theta_{k_0}^*), \Theta_k \Big\rangle }_{T_3}.
\end{aligned}
\end{equation}

The first term $T_1$ is just the usual statistical error term on $M$. For term $T_2$, we have
\begin{equation}
\label{eq:T2}
\sum_{i=1}^n \sum_{k=1}^K (T_2)^2 \leq \sum_{i=1}^n \sum_{k=1}^K \|E_i\|_F^2 \cdot \| \Theta_k^* - \Theta_k \|_F^2
= \bigg( \sum_{i=1}^n \|E_i\|_F^2 \bigg) \cdot \bigg(\sum_{k=1}^K  \| \Theta_k^* - \Theta_k \|_F^2 \bigg).
\end{equation}

For term $T_3$ we have
\begin{equation}
\label{eq:T3}
\begin{aligned}
\sum_{k=1}^K (T_3)^2 &\leq \bigg(\sum_{k=1}^K \|\Theta_k\|_F^2 \bigg)  \Big\| \sum_{k_0=1}^K m_{ik_0}^* (\Theta_{k_0} - \Theta_{k_0}^*) \Big\|_F^2 
\leq K\sigma_{\max}^2 \bigg(\sum_{k=1}^K  \| \Theta_k^* - \Theta_k \|_F^2 \bigg) \bigg( \sum_{k=1}^K {m_{ik}^*}^2 \bigg) \\
&\leq K\sigma_{\max}^2 \bigg(\sum_{k=1}^K  \| \Theta_k^* - \Theta_k \|_F^2 \bigg).
\end{aligned}
\end{equation}

Moreover, we have
\begin{equation}
\begin{aligned}
\| \Theta_k^* - \Theta_k \|_F &= \| {b_k^1}^*{b_k^2}^{*\top} - {b_k^1}{b_k^2}^{\top} \|_F \leq  \| {b_k^1}^*\|_2 \|{b_k^2}^* - b_k^2\|_2 + \| {b_k^2}\|_2 \|{b_k^1}^* - b_k^1\|_2 \\
&\leq 2 \sigma_{\max} \big( \|{b_k^2}^* - b_k^2\|_2 + \|{b_k^1}^* - b_k^1\|_2 \big),
\end{aligned}
\end{equation}
and hence
\begin{equation}
\begin{aligned}
\sum_{k=1}^K\| \Theta_k^* - \Theta_k \|_F^2 
\leq 4 {\sigma^2_{\max}} \sum_{k=1}^K \big( \|{b_k^2}^* - b_k^2\|_2 + \|{b_k^1}^* - b_k^1\|_2 \big)^2
\leq 8 {\sigma^2_{\max}} d^2(B,B^*).
\end{aligned}
\end{equation}

Combine \eqref{eq:T2} and \eqref{eq:T3}, taking summation over $i$, we obtain
\begin{equation}
\begin{aligned}
\frac{1}{n}\sum_{i=1}^n \sum_{k=1}^K (T_2)^2 + (T_3)^2 \leq 
\Big( e_0+K\sigma_{\max}^2 \Big) \cdot \bigg(\sum_{k=1}^K  \| \Theta_k^* - \Theta_k \|_F^2 \bigg)
\leq 8 {\sigma^2_{\max}} \Big( e_0+K\sigma_{\max}^2 \Big) \cdot d^2(B,B^*).
\end{aligned}
\end{equation}

\end{proof}

%% file: paper.bbl
\begin{thebibliography}{10}

\bibitem{Lauritzen1996Graphical}
{\em Graphical Models}, volume~17 of {\em Oxford Statistical Science Series}.
\newblock The Clarendon Press Oxford University Press, New York, 1996.
\newblock Oxford Science Publications.

\bibitem{ahmed2009recovering}
Amr Ahmed and Eric~P Xing.
\newblock Recovering time-varying networks of dependencies in social and
  biological studies.
\newblock {\em Proceedings of the National Academy of Sciences},
  106(29):11878--11883, 2009.

\bibitem{airoldi2013stochastic}
Edo~M Airoldi, Thiago~B Costa, and Stanley~H Chan.
\newblock Stochastic blockmodel approximation of a graphon: Theory and
  consistent estimation.
\newblock In {\em Advances in Neural Information Processing Systems}, pages
  692--700, 2013.

\bibitem{airoldi2008mixed}
Edoardo~M Airoldi, David~M Blei, Stephen~E Fienberg, and Eric~P Xing.
\newblock Mixed membership stochastic blockmodels.
\newblock {\em Journal of Machine Learning Research}, 9(Sep):1981--2014, 2008.

\bibitem{Barber2015ROCKET}
Rina~Foygel Barber and Mladen Kolar.
\newblock Rocket: Robust confidence intervals via kendall's tau for
  transelliptical graphical models.
\newblock {\em ArXiv e-prints, arXiv:1502.07641}, February 2015.

\bibitem{bickel2009nonparametric}
Peter~J Bickel and Aiyou Chen.
\newblock A nonparametric view of network models and newman--girvan and other
  modularities.
\newblock {\em Proceedings of the National Academy of Sciences},
  106(50):21068--21073, 2009.

\bibitem{burt2000network}
Ronald~S Burt.
\newblock The network structure of social capital.
\newblock {\em Research in organizational behavior}, 22:345--423, 2000.

\bibitem{chen2017fast}
Siheng Chen, Sufeng Niu, Leman Akoglu, Jelena Kova{\v{c}}evi{\'c}, and Christos
  Faloutsos.
\newblock Fast, warped graph embedding: Unifying framework and one-click
  algorithm.
\newblock {\em arXiv preprint arXiv:1702.05764}, 2017.

\bibitem{du2013uncover}
Nan Du, Le~Song, Hyenkyun Woo, and Hongyuan Zha.
\newblock Uncover topic-sensitive information diffusion networks.
\newblock In {\em Artificial Intelligence and Statistics}, pages 229--237,
  2013.

\bibitem{gao2018controllability}
Zuguang Gao, Xudong Chen, and Tamer Ba{\c{s}}ar.
\newblock Controllability of conjunctive boolean networks with application to
  gene regulation.
\newblock {\em IEEE Transactions on Control of Network Systems}, 5(2):770--781,
  2018.

\bibitem{gao2018stability}
Zuguang Gao, Xudong Chen, and Tamer Ba{\c{s}}ar.
\newblock Stability structures of conjunctive boolean networks.
\newblock {\em Automatica}, 89:8--20, 2018.

\bibitem{gehrke2003overview}
Johannes Gehrke, Paul Ginsparg, and Jon Kleinberg.
\newblock Overview of the 2003 kdd cup.
\newblock {\em ACM SIGKDD Explorations Newsletter}, 5(2):149--151, 2003.

\bibitem{gomez2010inferring}
Manuel Gomez~Rodriguez, Jure Leskovec, and Andreas Krause.
\newblock Inferring networks of diffusion and influence.
\newblock In {\em Proceedings of the 16th ACM SIGKDD international conference
  on Knowledge discovery and data mining}, pages 1019--1028. ACM, 2010.

\bibitem{gomez2015estimating}
Manuel Gomez-Rodriguez, Le~Song, Hadi Daneshmand, and Bernhard Sch{\"o}lkopf.
\newblock Estimating diffusion networks: Recovery conditions, sample complexity
  \& soft-thresholding algorithm.
\newblock {\em Journal of Machine Learning Research}, 2015.

\bibitem{hoff2009multiplicative}
Peter~D Hoff.
\newblock Multiplicative latent factor models for description and prediction of
  social networks.
\newblock {\em Computational and mathematical organization theory}, 15(4):261,
  2009.

\bibitem{hoff2002latent}
Peter~D Hoff, Adrian~E Raftery, and Mark~S Handcock.
\newblock Latent space approaches to social network analysis.
\newblock {\em Journal of the american Statistical association},
  97(460):1090--1098, 2002.

\bibitem{holland1983stochastic}
Paul~W Holland, Kathryn~Blackmond Laskey, and Samuel Leinhardt.
\newblock Stochastic blockmodels: First steps.
\newblock {\em Social networks}, 5(2):109--137, 1983.

\bibitem{hsu2016online}
Wei-Shou Hsu and Pascal Poupart.
\newblock Online bayesian moment matching for topic modeling with unknown
  number of topics.
\newblock In {\em NIPS}, 2016.

\bibitem{jarrah2010dynamics}
Abdul~Salam Jarrah, Reinhard Laubenbacher, and Alan Veliz-Cuba.
\newblock The dynamics of conjunctive and disjunctive boolean network models.
\newblock {\em Bulletin of Mathematical Biology}, 72(6):1425--1447, 2010.

\bibitem{jin2017estimating}
Jiashun Jin, Zheng~Tracy Ke, and Shengming Luo.
\newblock Estimating network memberships by simplex vertex hunting.
\newblock {\em arXiv preprint arXiv:1708.07852}, 2017.

\bibitem{karrer2010message}
Brian Karrer and Mark~EJ Newman.
\newblock Message passing approach for general epidemic models.
\newblock {\em Physical Review E}, 82(1):016101, 2010.

\bibitem{karrer2011stochastic}
Brian Karrer and Mark~EJ Newman.
\newblock Stochastic blockmodels and community structure in networks.
\newblock {\em Physical review E}, 83(1):016107, 2011.

\bibitem{kolar2010estimating}
Mladen Kolar, Le~Song, Amr Ahmed, and Eric~P Xing.
\newblock Estimating time-varying networks.
\newblock {\em The Annals of Applied Statistics}, pages 94--123, 2010.

\bibitem{kranton2001theory}
Rachel~E Kranton and Deborah~F Minehart.
\newblock A theory of buyer-seller networks.
\newblock {\em American economic review}, 91(3):485--508, 2001.

\bibitem{leskovec2005graphs}
Jure Leskovec, Jon Kleinberg, and Christos Faloutsos.
\newblock Graphs over time: densification laws, shrinking diameters and
  possible explanations.
\newblock In {\em Proceedings of the eleventh ACM SIGKDD international
  conference on Knowledge discovery in data mining}, pages 177--187. ACM, 2005.

\bibitem{nowicki2001estimation}
Krzysztof Nowicki and Tom A~B Snijders.
\newblock Estimation and prediction for stochastic blockstructures.
\newblock {\em Journal of the American statistical association},
  96(455):1077--1087, 2001.

\bibitem{parikh2014proximal}
Neal Parikh and Stephen Boyd.
\newblock Proximal algorithms.
\newblock {\em Foundations and Trends {\textregistered} in Optimization},
  1(3):127--239, 2014.

\bibitem{rodriguez2011uncovering}
Manuel~Gomez Rodriguez, David Balduzzi, and Bernhard Sch{\"o}lkopf.
\newblock Uncovering the temporal dynamics of diffusion networks.
\newblock {\em arXiv preprint arXiv:1105.0697}, 2011.

\bibitem{rohe2011spectral}
Karl Rohe, Sourav Chatterjee, and Bin Yu.
\newblock Spectral clustering and the high-dimensional stochastic blockmodel.
\newblock {\em The Annals of Statistics}, pages 1878--1915, 2011.

\bibitem{snijders1997estimation}
Tom~AB Snijders and Krzysztof Nowicki.
\newblock Estimation and prediction for stochastic blockmodels for graphs with
  latent block structure.
\newblock {\em Journal of classification}, 14(1):75--100, 1997.

\bibitem{walter2001value}
Achim Walter, Thomas Ritter, and Hans~Georg Gem{\"u}nden.
\newblock Value creation in buyer--seller relationships: Theoretical
  considerations and empirical results from a supplier's perspective.
\newblock {\em Industrial marketing management}, 30(4):365--377, 2001.

\bibitem{wang1987stochastic}
Yuchung~J Wang and George~Y Wong.
\newblock Stochastic blockmodels for directed graphs.
\newblock {\em Journal of the American Statistical Association}, 82(397):8--19,
  1987.

\bibitem{wasserman1994social}
Stanley Wasserman and Katherine Faust.
\newblock {\em Social network analysis: Methods and applications}, volume~8.
\newblock Cambridge university press, 1994.

\bibitem{wilson1995integrated}
David~T Wilson.
\newblock An integrated model of buyer-seller relationships.
\newblock {\em Journal of the academy of marketing science}, 23(4):335--345,
  1995.

\bibitem{Yu2017AnIM}
Ming Yu, Varun Gupta, and Mladen Kolar.
\newblock An influence-receptivity model for topic based information cascades.
\newblock {\em 2017 IEEE International Conference on Data Mining (ICDM)}, pages
  1141--1146, 2017.

\bibitem{yu2016statistical}
Ming Yu, Mladen Kolar, and Varun Gupta.
\newblock Statistical inference for pairwise graphical models using score
  matching.
\newblock In {\em Advances in Neural Information Processing Systems}, pages
  2829--2837, 2016.

\bibitem{yu2018recovery}
Ming Yu, Zhaoran Wang, Varun Gupta, and Mladen Kolar.
\newblock Recovery of simultaneous low rank and two-way sparse coefficient
  matrices, a nonconvex approach.
\newblock {\em arXiv preprint arXiv:1802.06967}, 2018.

\bibitem{zhou2013learning}
Ke~Zhou, Hongyuan Zha, and Le~Song.
\newblock Learning social infectivity in sparse low-rank networks using
  multi-dimensional hawkes processes.
\newblock In {\em Artificial Intelligence and Statistics}, pages 641--649,
  2013.

\end{thebibliography}
